\documentclass{article}

\usepackage[final]{rldm_2019}

\usepackage[utf8]{inputenc}
\usepackage[T1]{fontenc}    
\usepackage{hyperref}       
\usepackage{url}            
\usepackage{booktabs}       
\usepackage{amsfonts}       
\usepackage{nicefrac}       
\usepackage{microtype}      
\usepackage{amsmath}        
\usepackage{graphicx}

\RequirePackage{natbib}
\setcitestyle{authoryear,square,citesep={;},aysep={,},yysep={;}}
\RequirePackage{color}
\hypersetup{ 
    colorlinks=true,
    linkcolor=blue,
    citecolor=blue,
    filecolor=blue,
    urlcolor=blue,
    pdfview=FitH}

\title{Exploring Restart Distributions}

\author{
    Arash Tavakoli$^1$\thanks{Equal contribution. Correspondence to: 
    Arash Tavakoli (\texttt{\textcolor{blue}{a.tavakoli@imperial.ac.uk}}).} \\
    \And
    Vitaly Levdik$^{1*}$ \\
    \And
    Riashat Islam$^{2,3}$ \\
    \AND
    Christopher M. Smith$^{1}$ \\
    \And
    Petar Kormushev$^{1}$ \\
    \AND 
    $^1$\textnormal{Imperial College London,} 
    $^2$\textnormal{Mila,} 
    $^3$\textnormal{McGill University} 
}

\begin{document}

\maketitle

\begin{abstract}
We consider the generic approach of using an experience memory to help exploration by adapting a restart distribution. That is, given the capacity to reset the state with those corresponding to the agent's past observations, we help exploration by promoting faster state-space coverage via restarting the agent from a more diverse set of initial states, as well as allowing it to restart in states associated with significant past experiences. This approach is compatible with both on-policy and off-policy methods. However, a caveat is that altering the distribution of initial states could change the optimal policies when searching within a restricted class of policies. To reduce this unsought learning bias, we evaluate our approach in deep reinforcement learning which benefits from the high representational capacity of deep neural networks. We instantiate three variants of our approach, each inspired by an idea in the context of experience replay. Using these variants, we show that performance gains can be achieved, especially in hard exploration problems.
\end{abstract}

\section{Introduction}
\label{sec:intro}

Experience replay lets off-policy reinforcement learning (RL) methods remember and reuse past experiences \citep{lin1992experience, mnih2015human}. This helps circumvent the rapid forgetting of past experiences and, therefore, improves sample-efficiency. Prioritising experience can further boost efficiency by replaying important transitions more frequently, where different criteria may be considered to measure the importance of each transition. For example, the magnitude of a transition's temporal-difference (TD) error can be used as a proxy for how unexpected the transition is \citep{seijen2013planning, schaul2016prioritized}. Transitions can also be rated based on their corresponding episodic return \citep{oh2018self}, a particularly useful criterion in environments with sparse rewards.

On the other hand, on-policy methods cannot benefit from experience replay. As such, they are often sample-inefficient as past transitions are thrown away shortly after they are experienced, regardless of how rare or significant they may be. While replaying past experiences is not compatible with on-policy methods, creating new ones near previously-encountered states is. Given the capacity to reset the state with those corresponding to the agent's past observations (e.g. in a standard simulator), the latter can be made possible by maintaining a memory of the agent's previously-encountered states and using it to sample initial states. Effectively, this modifies the perceived distribution of initial states by combining the environment's \emph{initial-state distribution} with a proposal \emph{restart distribution} over the buffered states, where different criteria could be considered to \emph{prioritise} the latter.

We refer to this approach, generically, as \emph{exploring restart distributions}.\footnote{This choice was inspired in part by the theoretical assumption of \emph{exploring starts} \citep{sutton2018reinforcement}, with which our approach shares a subtle connection.} By drawing inspiration from well-known ideas in the context of experience replay, we instantiate three variants of our approach. Specifically, our \emph{uniform restart} resembles the uniform replay of \citep{mnih2015human}, our \emph{prioritised restart} resembles the prioritised replay of \citep{schaul2016prioritized}, and our \emph{episodic restart} resembles the episodic replay of \citep{oh2018self}. We combine our variants with a canonical policy-gradient algorithm, Proximal Policy Optimisation (PPO) \citep{schulman2017proximal}, which, due to its on-policy nature, cannot straightforwardly use experience replay and, as such, is interesting for our study. We test the resulting agents on two dense-reward and two sparse-reward environments, in each case considering a medium-difficulty and a hard exploration problem. We see improvements from our approach in all cases, with the most remarkable gains in the hard exploration problems.

Broadly, we consider \emph{simulator-based training} with \emph{simulator-free execution}, a problem paradigm in which we can utilise the opportunity to adjust certain environment variables during training but not during execution (e.g. \citep{ciosek2017offer}). Our approach improves this paradigm by utilising the reset capacity in simulated environments during training. We emphasise that we do not utilise this capacity during evaluations (i.e. policies are evaluated with respect to the original performance metric).\footnote{As such, our work differs from \citep{rajeswaran2017towards} which examines the impact of more diverse initial-state distributions in the context of ``robustness''.}

\section{Related work}
\label{sec:related_work}

\citet{kakade2002approximately} considered the notion of utilising the reset capacity and showed, under certain conditions, that using a proposal initial-state distribution that is more uniform over the state space than the original one improves learning performance with respect to the original performance metric. More recently, \citet{agarwal2019optimality} formalised the importance of how a favourable initial-state distribution provides a means to circumvent worst-case exploration issues in the context of policy-gradient methods \citep{sutton1999policy}. Nonetheless, these works do not provide a practical procedure for creating such distributions when the state space is unknown a priori.

To improve model-free learning, \citet{popov2017data} modified the initial-state distribution to be uniform over the states from provided expert demonstrations. \citet{salimans2018learning} reported high performance on Montezuma's Revenge Atari 2600 game by restarting a standard deep RL agent from a set of designated initial states, manually extracted from a single expert demonstration. These approaches resemble our episodic restart variant with one major difference: in our approach, the agent progressively updates its best buffered episodes in order to sample initial states from them and, as such, it does not rely on expert demonstrations or manually designated initial states.

\citet{ecoffet2019go} proposed a related method, called Go-Explore, that achieved the state-of-the-art on the hardest exploration games in the Atari 2600. Go-Explore's main principles are to maintain a memory of previously-encountered states, reset the environment to the ``promising'' ones to explore from, and repeat this process until a complete solution is found. Once found, a policy is trained by imitation learning on the solution trajectory. This work provides strong supporting evidence for the utility of exploration through restarting from previously-encountered states. However, Go-Explore does not use RL to learn a policy that solves the problem. Furthermore, the criterion used to identify promising states is rather domain-specific. Using our approach, one could realise an RL counterpart for Go-Explore by using the same criterion to prioritise initial states. 

\citet{florensa2017reverse} presented a method for adaptive generation of curricula in the form of initial-state distributions that start close to the goal state and gradually move away with the agent's progress. This method is limited to goal-oriented problems with clear goal states and further assumes a priori knowledge of such states. While our approach is not limited to such environments, a similar behaviour to curriculum generation in this way could emerge with our approach using an appropriate priority measure, whereby a single encounter of a goal state biases the restart distribution towards it.

Restarting from previously-encountered states to sample more transitions reduces the variance of the gradient estimator in policy-gradient methods. The vine procedure of \citep{schulman2015trust} utilises the reset capacity in simulated environments for this purpose. This method can be realised as a special case of our approach.

\section{Background}
\label{sec:background}

We consider the RL framework \citep{szepesvari2010algorithms, sutton2018reinforcement} in which the interaction of an agent and an environment is modeled as a Markov decision process (MDP) \citep{puterman1994markov} comprising of a state space $\mathcal{S}$, an action space $\mathcal{A}$, an initial-state distribution $p_1(s_1) = Pr\{S_1 {=} s_1\}$, a transition distribution $p(s'|s,a) = Pr\{S_{t+1} {=} s'|S_t {=} s,A_t {=} a\}$, and a reward function $r(s,a,s') = \mathbb{E}[R_{t}|S_t {=} s,A_t {=} a,S_{t+1} {=} s']$, for all $s,s' \in \mathcal{S}, a \in \mathcal{A}, s_1 \in \mathcal{S}_1 \subset \mathcal{S}$. The decision-making process of an agent is characterised by a policy $\pi(a|s) = Pr\{A_t {=} a|S_t {=} s\}$. This policy can be approximated by a parameterised function $\pi(a|s,{\boldsymbol \theta})$ (e.g. a neural network), where ${\boldsymbol \theta} \in \mathbb{R}^d$ is the vector of policy parameters and, typically, $d \ll |\mathcal{S}|$. The agent uses its policy to interact with the environment to sample a trajectory $S_1, A_1, R_1, S_2, \dots, S_T, A_T, R_T, S_{T+1}$ (where $T$ is the trajectory's horizon which is, in general, a random variable). In this paper, we assume that $T$ is finite and that terminations may occur due to terminal states in episodic tasks (i.e. \emph{concrete episodes}) or due to an arbitrary condition, such as timeouts, in continuing or episodic tasks (i.e. \emph{partial episodes}). The majority of our discussions are considered under the more generic assumption of learning from partial episodes and, as such, are relevant only to bootstrapping methods \citep{sutton2018reinforcement} (e.g. TD methods such as Q-learning, Sarsa, and actor-critic methods). Nevertheless, the main proposition of this paper applies also to Monte-Carlo methods, in which case the episodes are strictly concrete. 

We assume access to the capacity to reset the state with those corresponding to the agent’s past observations. We remark that this assumption is weaker than having explicit access to the environment's model. Furthermore, we do not assume a priori knowledge of the (valid) state space. In fact, such knowledge is rarely accessible in practice, which is why we build a memory of states on-the-fly.

\subsection{Impact of the initial-state distribution on the learning objective}
\label{sec:bias}

In this section, we consider the question ``how does modifying the initial-state distribution affect the learning objective and, ultimately, the learned policy with respect to the original performance metric?''. We will consider this question separately for tabular and approximate solution methods.

In tabular methods, the learned values at each state are decoupled from one another (i.e. an update at one state affects no other). Let us now consider the control problem in which the agent's goal is to maximise its value from the environment's designated set of initial states. As per the \emph{principle of optimality} \citep{sutton2018reinforcement}, a policy achieves the optimal value from a state $s$, if and only if, for any state $s'$ reachable from $s$ it achieves the optimal value. Therefore, by letting the agent also start in states outside the environment's designated set of initial states, we can better optimise for the designated set by better optimising for the states that are reachable from the designated set.

On the contrary, with approximation, an update at one state affects many others as generally we have far more states than parameters. Therefore, making one state's estimate more accurate often means making others' less accurate. Let us now consider a common objective function for approximate prediction of the action values for a given policy:
\begin{equation}
    L(\mathbf{w}) \doteq 
    \sum_{s\in \mathcal{S}} \textcolor{blue}{\rho(s)} \sum_{a\in \mathcal{A}} \pi(a|s) \Big(q_\pi(s,a) - \hat{q}_\pi(s,a|\mathbf{w})\Big)^2.    
    \label{equ:prediction}
\end{equation}
This objective function is weighted according to the state distribution $\textcolor{blue}{\rho(s)}$, which depends on the policy $\pi(a|s)$ and, in episodic tasks, the initial-state distribution $p_1(s_1)$. In effect, this promotes the approximation of the action values to be more accurate at states that have a higher visitation density. As such, changing the initial-state distribution modifies the learning objective for approximate prediction. The same rationale holds in approximate control, e.g. when using policy-gradient methods. One caveat to this in the control case is a policy that maximises the modified learning objective within some restricted class of policies may perform poorly with respect to the original performance metric. We can reduce this unsought learning bias by using a distribution of initial states whose support contains and spans beyond that of the environment's initial-state distribution (see Sec.~\ref{sec:method}), as well as using a parameterisation that affords the problem's underlying complexity. While the latter cannot be guaranteed in general, it seems often admissible in deep RL (especially considering the relative simplicity of many problems of interest with respect to the commonly-used, high-capacity neural networks \citep{rajeswaran2017towards}). Nonetheless, there are many problems where learning any reasonable policy is challenging, not to mention learning an optimal one. In such cases, it may be appropriate to accept the cost of potentially introducing a learning bias in order to facilitate learning.

\section{Exploring restart distributions}
\label{sec:method}

We consider a generic approach in which the agent maintains, what we call, a \emph{restart memory} of its past experiences along with their corresponding (true) states, and uses this restart memory to sample initial states for new episodes. This in turn allows the agent to gradually increase the diversity of the states in which it can restart. Formally, for the environment's initial-state space $\mathcal{S}_1$ and the agent's set of buffered states $\mathcal{S}_\mathcal{B}$, our approach enables sampling initial states from $\mathcal{S}_1 \cup \mathcal{S}_\mathcal{B}$ (which contains and spans beyond $\mathcal{S}_1$). We achieve this by sampling from both the environment's initial-state distribution $p_1$ (with support $\mathcal{S}_1$) and a restart distribution $p_\mathcal{B}$ (with support $\mathcal{S}_\mathcal{B}$). This is equivalent to sampling from a new distribution $\mu_1$, which mixes $p_1$ and $p_\mathcal{B}$, with support $\mathcal{S}_1 \cup \mathcal{S}_\mathcal{B}$. In this paper, we control the extent of contributions from each of the distributions $p_1$ and $p_\mathcal{B}$ by maintaining a fixed \emph{ratio} for the number of transitions that stem from \emph{augmented initial states} (i.e. initial states that are sampled from a restart memory) versus the total number of transitions.

\subsection{Uniform restart} 
\label{sec:uniform_restart}

Our uniform restart variant generally follows the same mechanism as the uniform replay of \citep{mnih2015human}. The differences are that we store states as opposed to observations and that we ultimately care about sampling states as opposed to transitions. In other words, we store recent states (without selection) in a restart memory $\mathcal{B}$ and sample augmented initial states uniformly (i.e. $p_\mathcal{B}$ is a uniform distribution over the buffered states $\mathcal{S}_\mathcal{B}$).

Having the capacity to reset the state naturally implies we can early-terminate episodes. Doing so is compatible with bootstrapping methods which can bootstrap at the end of partial episodes \citep{pardo2018time}. By convention, we choose to apply a time limit $T_{\textrm{aug}}$ to interactions that stem from augmented initial states with $T_{\textrm{aug}} \leq T_{\textrm{env}}$, where $T_{\textrm{env}}$ is the environment's time limit (if any).

\subsection{Prioritised restart} 
\label{sec:prior_restart}

Our prioritised restart variant uses a similar mechanism as the \emph{proportional} prioritised replay of \citep{schaul2016prioritized} but for prioritising states rather than transitions. As such, we use the state-value TD error (as opposed to the state-action form used in \citep{schaul2016prioritized}):
\begin{equation}
    \delta_i \doteq r_i + \gamma v(s'_i) - v(s_i) \,,
\label{equ:state_TD}
\end{equation}
where $i$ is the index of state $s_i$ in the restart memory. We calculate the probability of sampling state $s_i$ from the restart memory via
\begin{equation}
    p_\mathcal{B}(s_i) = \frac{p_i^\alpha}{\sum_k{p_k^\alpha}} \,,
\label{equ:prioritised_prob}
\end{equation}
where $p_i \doteq |\delta_i| + \varepsilon$ is the priority of state $s_i$ (with $\varepsilon$ being a small positive constant to ensure non-zero probabilities for all buffered states) and the exponent $\alpha$ determines how much prioritisation is used. It is noteworthy that prioritising initial states does not introduce a learning bias in the way that prioritisation in the context of replay does. Biasing the replay frequency directly alters the perceived state transition and reward dynamics in stochastic environments. However, this does not apply to our approach as, regardless of how an initial state is sampled, transitions are always sampled from the environment, not replayed from a replay memory. Therefore, we do not need importance sampling corrections as used in \citep{schaul2016prioritized}. Lastly, similar to our uniform restart variant, we apply a time limit $T_{\textrm{aug}}$ to interactions that stem from augmented initial states.

\subsection{Episodic restart} 
\label{sec:episodic_restart}

Episodic return is another criterion for measuring priorities, one that is particularly useful in environments with sparse rewards \citep{oh2018self}. We build our episodic restart variant to enable using this criterion for prioritisation, leading to a number of differences with respect to our previous variants. Most fundamentally, states are now buffered at the end of episodes rather than on each transition, and only episodes that obtain a higher undiscounted return in comparison to those already in the restart memory are buffered. In other words, the agent maintains its most rewarding episodes, where the restart-memory size determines the maximum number of episodes in the restart memory at any given time. The episodes are then prioritised according to their corresponding undiscounted returns, with uniform sampling of states from any selected episodes. While it is possible to also prioritise the states in a selected episode (e.g. using TD error), we omit that in this work to simplify our experiments.

Using our episodic restart variant in the manner described above could significantly hurt the agent's learning performance by biasing its experiences towards specific parts of the state space. For example, consider a multi-goal problem in which the goal is different in every episode and, as such, the goal is part of the state (i.e. assuming \emph{Markov states} \citep{sutton2018reinforcement}). Encountering an episode that has an easy goal could result in obtaining a high return in comparison to other episodes.
The states from such an episode receive a high priority for being sampled as initial states in new episodes. This leads to frequently experiencing episodes of the same easy goal, thereby quickly filling the restart memory with episodes of a single goal. To address this, we use a nested storage mechanism in which an episode that stems from an environment's initial state is stored as a \emph{parent episode} (i.e. a standard episode) and an episode that stems from an augmented initial state is stored as a \emph{sub-episode}, where each sub-episode is linked to the parent episode from which its augmented initial state originated. This avoids filling the restart memory with episodes of similar nature as, now, only a limited number of sub-episodes can be stored under each parent episode, with the nature of each parent episode being determined by the environment's initial-state distribution. We will now discuss what is needed and how to sample an augmented initial state from this restart memory.

To maintain episodic returns comparable in environments with time limits, all sub-episodes need to be terminated after an appropriate number of steps, $T_{\textrm{aug}}$. Assuming (as before) a fixed, 
environment's time limit $T_{\textrm{env}}$, sub-episodic time limit $T_{\textrm{aug}}$ can be determined via 
\begin{equation}
\label{equ:aug-time}
    T_{\textrm{aug}} = T_{\textrm{env}}-t \;\; \textrm{with} \;\; t \leq T_{\textrm{env}} - 1, 
\end{equation}
where $t$ is the time step of the augmented initial state.\footnote{The number of steps in the shortest path from the augmented initial state to the initial state of its corresponding parent episode.} In this way, we can calculate an augmented episodic return for any sub-episodes as the sum of the sub-episode's rewards and the rewards along the shortest path from the parent episode's initial state to the sub-episode's augmented initial state. 

To calculate the priority of a \emph{category of episodes} (i.e. a parent episode together with its sub-episodes), we use the maximum undiscounted return across the category's parent episode and all its sub-episodes. Let us denote this maximum undiscounted return per category with $\bar{G}_i$, with $i$ being the index of the category in the restart memory. We calculate the priority of the $i$th category as
\begin{equation}
    p_i \doteq \bar{G}_i - \delta_i + \varepsilon \,,
\label{equ:episodic-prob-1}
\end{equation}
where
\begin{equation}
    \delta_i \doteq \min(0, \min_{i} \bar{G}_i)
\label{equ:episodic-prob-2}
\end{equation}
is used as an offset to enable handling negative returns and $\varepsilon$ is a small positive constant that ensures non-zero probabilities for all buffered categories. 
We calculate the probability of sampling the $i$th category from the restart memory by using the priorities of Eq.~(\ref{equ:episodic-prob-1}) in Eq.~(\ref{equ:prioritised_prob}). We perform this once to sample a category of episodes, and again to sample an episode from the selected category.\footnote{Each sub-episode in a category of episodes is augmented with as many as $T_{\textrm{aug}}$ states which link its augmented initial state to the initial state of its corresponding parent episode.} We then sample an augmented initial state uniformly from the selected episode.

\section{Experiments}
\label{sec:experiments}

We evaluate our approach on several continuous control environments, simulated using the MuJoCo physics engine \citep{todorov2012mujoco}.
As per the nature of our variants, we present performance evaluations of our uniform and prioritised restart variants in dense-reward environments (Sec.~\ref{sec:dense_reward_results}) and of our episodic restart variant in sparse-reward environments (Sec.~\ref{sec:sparse_reward_results}). In each case, we consider a medium-difficulty and a hard exploration problem.

We focus our experiments on on-policy RL which cannot straightforwardly replay past experiences and, thus, will benefit more significantly from our approach. Specifically, we evaluate our approach using PPO, which is a canonical on-policy method for continuous control. To avoid the significant cost of systematic hyperparameter search, throughout our experiments we fix the ratio hyperparameter of our approach to 0.1 (i.e. 10\% of the total interactions stem from augmented initial states) and, generally, use the PPO hyperparameters as originally reported in \citep{schulman2017proximal}.

\subsection{Dense-reward environments}  
\label{sec:dense_reward_results}

To evaluate the performance of our uniform and prioritised restart variants, we consider two dense-reward environments from the OpenAI Gym \citep{brockman2016openai}, namely HalfCheetah (medium-difficulty exploration) and Humanoid (hard exploration). By default, these environments apply a time limit of $T_{\textrm{env}} = \textrm{1000}$ to each episode. We apply this default time limit to all interactions that stem from the environment's initial states. For any interactions that stem from an augmented initial state, we apply the much shorter time limit of $T_{\textrm{aug}} = \textrm{10}$. For both of our variants, we set the restart-memory size to 20000. For our prioritised restart variant, we set $\alpha = \textrm{0.4}$ to induce a mild priority (see Eq.~(\ref{equ:prioritised_prob})).

\subsubsection{HalfCheetah}
\label{sec:halfcheetah}

The goal in the HalfCheetah environment is to make a planar biped run as fast as possible. Given the medium-dimensionality of its observation and action spaces, this environment is not very challenging for advanced RL agents. Fig.~\ref{fig:dense_reward_results} (left) shows the learning curves for this experiment, created by evaluating each agent periodically during training using the environment's initial-state distribution. The learning curves show average undiscounted returns (each averaged over 5 seeds). Our results show mild improvements for both of our variants, with a slight advantage for our prioritised one.

\subsubsection{Humanoid}
\label{sec:humanoid}

\begin{figure}[t!]
  \centering
  \includegraphics[height=1.9in]{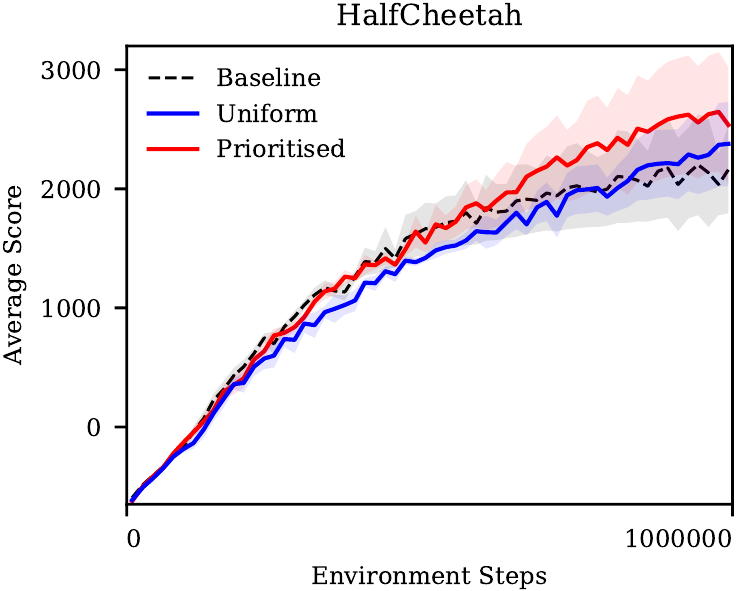}
  \hspace{.05in}
  \includegraphics[height=1.9in]{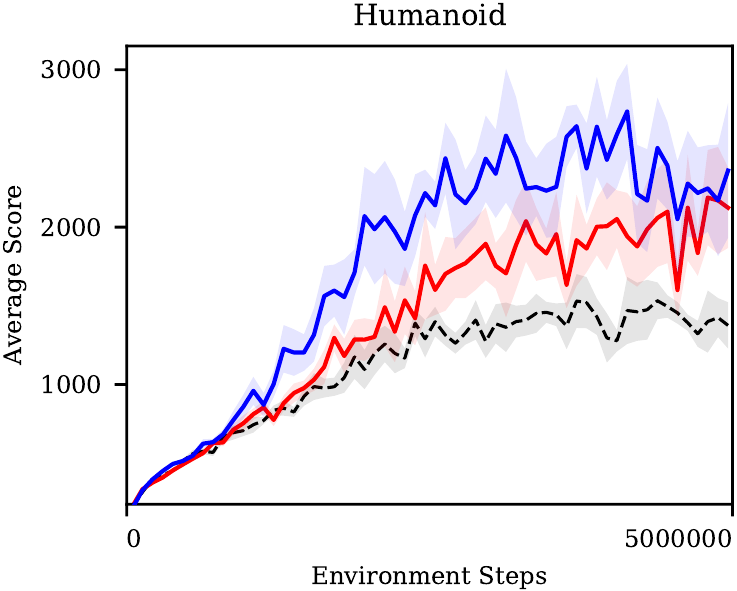}
  \caption{Average test performance curves of our uniform and prioritised restart variants as applied to and against PPO on two dense-reward environments. Shaded areas are standard error.}
  \label{fig:dense_reward_results}
\end{figure}

The goal in the Humanoid environment is to make a three-dimensional biped walk forward as fast as possible, without falling over. While a dense-reward environment, the high-dimensionality of its observation and action spaces make it rather challenging. Fig.~\ref{fig:dense_reward_results} (right) shows the learning curves for this experiment, created following the same procedure as in our HalfCheetah experiment. Our results show significant improvements for the agents that use our uniform or prioritised restart variants. However, contrary to our HalfCheetah results, here our uniform restart variant outperforms our prioritised one. We conducted an informal study to examine whether one of these variants would have a general advantage but did not find the results to be conclusive in this regard.

\subsection{Sparse-reward environments}
\label{sec:sparse_reward_results}

To evaluate the performance of our episodic restart variant, we consider two sparse-reward environments from the OpenAI Gym, namely FetchReach (medium-difficulty exploration) and Thrower (hard exploration).

\subsubsection{FetchReach}
\label{sec:fetchreach}

The FetchReach environment was proposed by \citet{plappert2018multi} to assess goal-conditional policy learning methods in a problem of practical interest. In each episode, the agent's task is to control its four joints in order to move the gripper to a goal position, where the goal is randomly sampled at the start of each episode. The agent's observations consist of the robot's current state as well as a three-dimensional goal position. This environment has no terminal states but enforces a time limit of $T_{\textrm{env}} = \textrm{50}$. We set the restart-memory size to 100 parent episodes and 10 sub-episodes. The agent receives a step-wise penalty of -1 whenever its gripper is not at the goal position, and 0 otherwise.

\begin{figure}[t!]
  \centering
  \includegraphics[height=1.9in]{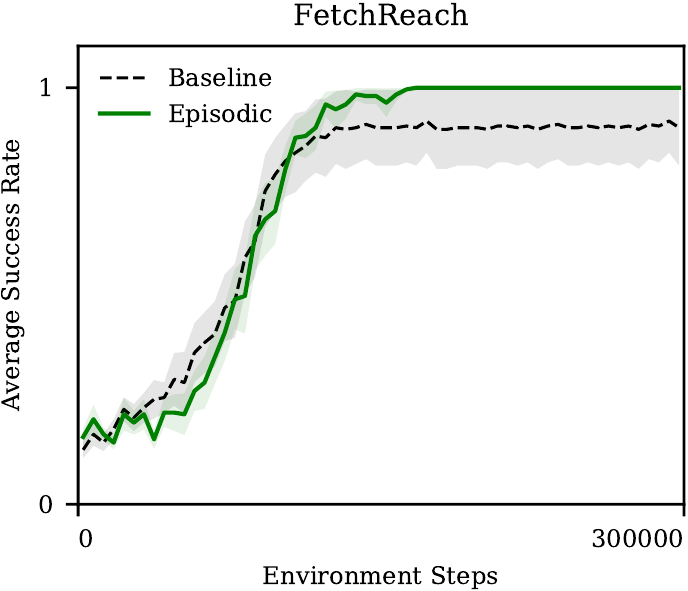}
  \hspace{.05in}
  \includegraphics[height=1.9in]{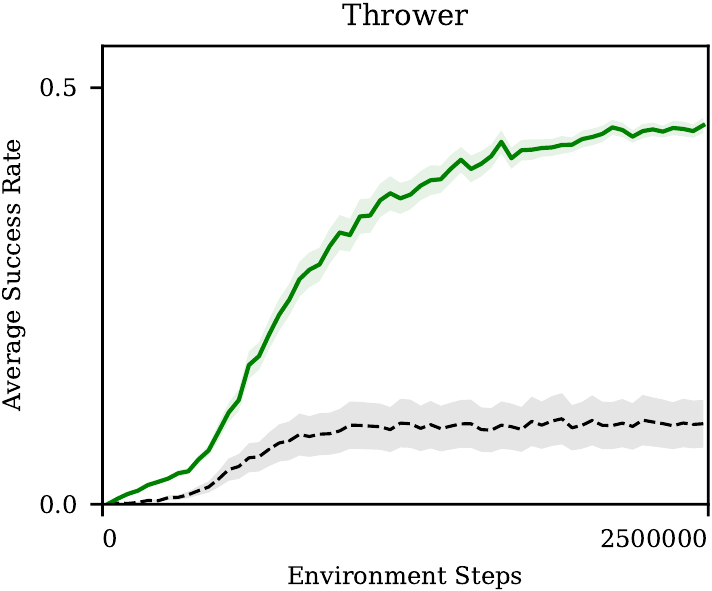}
  \caption{Average test success-rate curves of our episodic restart variant as applied to and against PPO on two sparse-reward environments. Shaded areas are standard error.}
  \label{fig:sparse_reward_results}
\end{figure}

Fig.~\ref{fig:sparse_reward_results} (left) shows the learning curves for this experiment, created by evaluating each agent periodically during training using the environment's initial-state distribution.
The learning curves show average success rates (each averaged over 10 seeds). Our results show mild improvements for our episodic restart variant. This is consistent with our observation in the HalfCheetah experiment, suggesting that performance improvements can be expected from our approach even in environments which do not pose a significant exploration challenge.

\subsubsection{Thrower}
\label{sec:thrower}

Our Thrower environment is a variant of the one provided in the OpenAI Gym, modified to feature significantly sparse rewards. Our modifications make this problem particularly challenging as the agent receives a positive reward of 1 only for successfully throwing the ball in the goal region, and 0 otherwise. The agent additionally incurs a small torque penalty. Each episode terminates once the ball impacts the goal or the table, or upon reaching the time limit of $T_{\textrm{env}} = \textrm{100}$. We set the restart-memory size to 50 parent episodes and 10 sub-episodes. Due to the complexity of this environment, the probability of the ball impacting the goal is very low. Hence, to ensure that each independent run encounters a positive training signal early on in its training process, we only considered runs that experienced an instance of the ball impacting the goal in the first 50000 interaction steps (roughly 10\% of our runs achieved this during the specified window). Moreover, we use an entropy coefficient of 0.02 to further encourage exploration.

Fig.~\ref{fig:sparse_reward_results} (right) shows the learning curves for this experiment, created following the same procedure as in our FetchReach experiment (each averaged over 42 seeds). Our results show a clear advantage for using our episodic restart variant. By inspecting the performances of individual runs, we found that our episodic restart variant enabled the agent to learn consistently across all 42 independent runs (with each run achieving just under 50\% success rate per evaluation trial), whereas the baseline agent completely failed to learn any useful policy in 80\% of the runs (with the remaining runs achieving just under 50\% success rate per evaluation trial, similar to our variant). In other words, using our episodic restart variant led to more robust learning (with respect to random initialisation) by enabling a way for the agent to utilise its extremely rare positive experiences.

\section{Conclusion}
\label{sec:conclusion}

We considered the generic approach of maintaining a restart memory of the agent's past experiences along with their corresponding (true) states, and using this restart memory to sample initial states for new episodes. This approach utilises the reset capacity in simulated environments during training in order to help with exploration. We instantiated three variants of our approach by drawing inspiration from well-known ideas in the context of experience replay. We tested these variants on two dense-reward and two sparse-reward environments. In each case, we considered a medium-difficulty and a hard exploration problem. We showed improvements from our approach in all cases, with the most remarkable gains in the hard exploration problems.

\subsubsection*{Acknowledgements}

AT acknowledges financial support from the UK Engineering and Physical Sciences Research Council (EPSRC DTP). VL and PK acknowledge financial support from the Samsung Advanced Institute of Technology (SAIT GRO). All authors acknowledge computational resources from Microsoft (Azure for Research Award).

\bibliography{references}

\begin{thebibliography}{23}
\providecommand{\natexlab}[1]{#1}
\providecommand{\url}[1]{\texttt{#1}}
\expandafter\ifx\csname urlstyle\endcsname\relax
  \providecommand{\doi}[1]{doi: #1}\else
  \providecommand{\doi}{doi: \begingroup \urlstyle{rm}\Url}\fi

\bibitem[Lin(1992)]{lin1992experience}
Long-Ji Lin.
\newblock Self-improving reactive agents based on reinforcement learning,
  planning and teaching.
\newblock \emph{Machine Learning}, 8\penalty0 (3):\penalty0 293--321, 1992.

\bibitem[Mnih et~al.(2015)Mnih, Kavukcuoglu, Silver, Rusu, Veness, Bellemare,
  Graves, Riedmiller, Fidjeland, Ostrovski, Petersen, Beattie, Sadik,
  Antonoglou, King, Kumaran, Wierstra, Legg, and Hassabis]{mnih2015human}
Volodymyr Mnih, Koray Kavukcuoglu, David Silver, Andrei~A. Rusu, Joel Veness,
  Marc~G. Bellemare, Alex Graves, Martin Riedmiller, Andreas~K. Fidjeland,
  Georg Ostrovski, Stig Petersen, Charles Beattie, Amir Sadik, Ioannis
  Antonoglou, Helen King, Dharshan Kumaran, Daan Wierstra, Shane Legg, and
  Demis Hassabis.
\newblock Human-level control through deep reinforcement learning.
\newblock \emph{Nature}, 518\penalty0 (7540):\penalty0 529--533, 2015.

\bibitem[van Seijen and Sutton(2013)]{seijen2013planning}
Harm van Seijen and Richard~S. Sutton.
\newblock Planning by prioritized sweeping with small backups.
\newblock In \emph{Proceedings of the International Conference on Machine
  Learning}, pages 361--369, 2013.

\bibitem[Schaul et~al.(2016)Schaul, Quan, Antonoglou, and
  Silver]{schaul2016prioritized}
Tom Schaul, John Quan, Ioannis Antonoglou, and David Silver.
\newblock Prioritized experience replay.
\newblock In \emph{Proceedings of the International Conference on Learning
  Representations}, 2016.

\bibitem[Oh et~al.(2018)Oh, Guo, Singh, and Lee]{oh2018self}
Junhyuk Oh, Yijie Guo, Satinder~P. Singh, and Honglak Lee.
\newblock Self-imitation learning.
\newblock In \emph{Proceedings of the International Conference on Machine
  Learning}, pages 3878--3887, 2018.

\bibitem[Sutton and Barto(2018)]{sutton2018reinforcement}
Richard~S. Sutton and Andrew~G. Barto.
\newblock \emph{Reinforcement Learning: {A}n Introduction}.
\newblock MIT {P}ress, 2nd edition, 2018.

\bibitem[Schulman et~al.(2017)Schulman, Wolski, Dhariwal, Radford, and
  Klimov]{schulman2017proximal}
John Schulman, Filip Wolski, Prafulla Dhariwal, Alec Radford, and Oleg Klimov.
\newblock Proximal policy optimization algorithms.
\newblock \emph{arXiv:1707.06347}, 2017.

\bibitem[Ciosek and Whiteson(2017)]{ciosek2017offer}
Kamil Ciosek and Shimon Whiteson.
\newblock {OFFER}: {O}ff-environment reinforcement learning.
\newblock In \emph{Proceedings of the AAAI Conference on Artificial
  Intelligence}, pages 1819--1825, 2017.

\bibitem[Rajeswaran et~al.(2017)Rajeswaran, Lowrey, Todorov, and
  Kakade]{rajeswaran2017towards}
Aravind Rajeswaran, Kendall Lowrey, Emanuel Todorov, and Sham~M. Kakade.
\newblock Towards generalization and simplicity in continuous control.
\newblock In \emph{Advances in Neural Information Processing Systems}, pages
  6550--6561, 2017.

\bibitem[Kakade and Langford(2002)]{kakade2002approximately}
Sham~M. Kakade and John Langford.
\newblock Approximately optimal approximate reinforcement learning.
\newblock In \emph{Proceedings of the International Conference on Machine
  Learning}, pages 267--274, 2002.

\bibitem[Agarwal et~al.(2019)Agarwal, Kakade, Lee, and
  Mahajan]{agarwal2019optimality}
Alekh Agarwal, Sham~M. Kakade, Jason~D. Lee, and Gaurav Mahajan.
\newblock Optimality and approximation with policy gradient methods in {M}arkov
  decision processes.
\newblock \emph{arXiv:1908.00261}, 2019.

\bibitem[Sutton et~al.(1999)Sutton, McAllester, Singh, and
  Mansour]{sutton1999policy}
Richard~S. Sutton, David~A. McAllester, Satinder~P. Singh, and Yishay Mansour.
\newblock Policy gradient methods for reinforcement learning with function
  approximation.
\newblock In \emph{Advances in Neural Information Processing Systems}, pages
  1057--1063, 1999.

\bibitem[Popov et~al.(2017)Popov, Heess, Lillicrap, Hafner, Barth-Maron,
  Vecerik, Lampe, Tassa, Erez, and Riedmiller]{popov2017data}
Ivaylo Popov, Nicolas Heess, Timothy~P. Lillicrap, Roland Hafner, Gabriel
  Barth-Maron, Matej Vecerik, Thomas Lampe, Yuval Tassa, Tom Erez, and Martin
  Riedmiller.
\newblock Data-efficient deep reinforcement learning for dexterous
  manipulation.
\newblock \emph{arXiv:1704.03073}, 2017.

\bibitem[Salimans and Chen(2018)]{salimans2018learning}
Tim Salimans and Richard Chen.
\newblock Learning {M}ontezuma's {R}evenge from a single demonstration.
\newblock \emph{arXiv:1812.03381}, 2018.

\bibitem[Ecoffet et~al.(2019)Ecoffet, Huizinga, Lehman, Stanley, and
  Clune]{ecoffet2019go}
Adrien Ecoffet, Joost Huizinga, Joel Lehman, Kenneth~O. Stanley, and Jeff
  Clune.
\newblock Go-{E}xplore: {A} new approach for hard-exploration problems.
\newblock \emph{arXiv:1901.10995}, 2019.

\bibitem[Florensa et~al.(2017)Florensa, Held, Wulfmeier, Zhang, and
  Abbeel]{florensa2017reverse}
Carlos Florensa, David Held, Markus Wulfmeier, Michael Zhang, and Pieter
  Abbeel.
\newblock Reverse curriculum generation for reinforcement learning.
\newblock In \emph{Proceedings of the Conference on Robot Learning}, pages
  482--495, 2017.

\bibitem[Schulman et~al.(2015)Schulman, Levine, Abbeel, Jordan, and
  Moritz]{schulman2015trust}
John Schulman, Sergey Levine, Pieter Abbeel, Michael Jordan, and Philipp
  Moritz.
\newblock Trust region policy optimization.
\newblock In \emph{Proceedings of the International conference on machine
  learning}, pages 1889--1897, 2015.

\bibitem[Szepesv\'ari(2010)]{szepesvari2010algorithms}
Csaba Szepesv\'ari.
\newblock \emph{Algorithms for Reinforcement Learning}.
\newblock Morgan \& Claypool, 2010.

\bibitem[Puterman(1994)]{puterman1994markov}
Martin~L. Puterman.
\newblock \emph{Markov Decision Processes: {D}iscrete Stochastic Dynamic
  Programming}.
\newblock John Wiley \& Sons, 1994.

\bibitem[Pardo et~al.(2018)Pardo, Tavakoli, Levdik, and
  Kormushev]{pardo2018time}
Fabio Pardo, Arash Tavakoli, Vitaly Levdik, and Petar Kormushev.
\newblock Time limits in reinforcement learning.
\newblock In \emph{Proceedings of the International Conference on Machine
  Learning}, pages 4045--4054, 2018.

\bibitem[Todorov et~al.(2012)Todorov, Erez, and Tassa]{todorov2012mujoco}
Emanuel Todorov, Tom Erez, and Yuval Tassa.
\newblock {MuJoCo}: {A} physics engine for model-based control.
\newblock In \emph{Proceedings of the IEEE/RSJ International Conference on
  Intelligent Robots and Systems}, pages 5026--5033, 2012.

\bibitem[Brockman et~al.(2016)Brockman, Cheung, Pettersson, Schneider,
  Schulman, Tang, and Zaremba]{brockman2016openai}
Greg Brockman, Vicki Cheung, Ludwig Pettersson, Jonas Schneider, John Schulman,
  Jie Tang, and Wojciech Zaremba.
\newblock {OpenAI Gym}.
\newblock \emph{arXiv:1606.01540}, 2016.

\bibitem[Plappert et~al.(2018)Plappert, Andrychowicz, Ray, McGrew, Baker,
  Powell, Schneider, Tobin, Chociej, Welinder, Kumar, and
  Zaremba]{plappert2018multi}
Matthias Plappert, Marcin Andrychowicz, Alex Ray, Bob McGrew, Bowen Baker,
  Glenn Powell, Jonas Schneider, Josh Tobin, Maciek Chociej, Peter Welinder,
  Vikash Kumar, and Wojciech Zaremba.
\newblock Multi-goal reinforcement learning: {C}hallenging robotics
  environments and request for research.
\newblock \emph{arXiv:1802.09464}, 2018.

\end{thebibliography}
\bibliographystyle{unsrtnat}

\end{document}